\documentclass[runningheads, envcountsame, a4paper]{llncs}

\usepackage{microtype}
\usepackage{graphicx}
\usepackage{subfigure}
\usepackage{booktabs}
\usepackage{amssymb}
\usepackage{amsmath}
\usepackage{amsfonts}       
\usepackage{nicefrac}       
\usepackage{lipsum}
\usepackage[hyphens]{url}            
\usepackage{hyperref}
\usepackage[linesnumbered,algoruled,boxed,lined]{algorithm2e}
\usepackage{bbold}
\usepackage{multirow}
\usepackage{tabularx}
\usepackage{hyphenat}
\usepackage[misc]{ifsym}

\DontPrintSemicolon

\begin{document}
\title{Fairness by Explicability and Adversarial SHAP Learning\thanks{Supported by Experian Ltd.}}
\toctitle{Fairness by Explicability and Adversarial SHAP Learning}

\author{James M. Hickey(\Letter)\and Pietro G. Di Stefano\and Vlasios Vasileiou}
\authorrunning{J. Hickey, P. Di Stefano, and V. Vasileiou}
\tocauthor{James M. Hickey (Experian UK\&I and EMEA DataLabs),
Pietro G. Di Stefano (Experian UK\&I and EMEA DataLabs),
Vlasios Vasileiou (Experian UK\&I and EMEA DataLabs),
}

\institute{Experian UK\&I and EMEA DataLabs\\London\\UK\\
\email{james.hickey@experian.com}}
\maketitle
\setcounter{footnote}{0}

\begin{abstract}
The ability to understand and trust the fairness of model predictions, particularly when considering the outcomes of unprivileged groups, is critical to the deployment and adoption of machine learning systems. SHAP values provide a unified framework for interpreting model predictions and feature attribution but do not address the problem of fairness directly. In this work, we propose a new definition of fairness that emphasises the role of an external auditor and model explicability. To satisfy this definition, we develop a framework for mitigating model bias using regularizations constructed from the SHAP values of an adversarial surrogate model. We focus on the binary classification task with a single unprivileged group and link our fairness explicability constraints to classical statistical fairness metrics. We demonstrate our approaches using gradient and adaptive boosting on: a synthetic dataset, the UCI Adult (Census) dataset and a real-world credit scoring dataset. The models produced were fairer and performant.
\end{abstract}

\section{Introduction}
\label{intro}
The last few decades have seen machine learning algorithms become even more performant and leverage larger varieties of data. These advances have led to wide-spread adoption of machine learning in nearly every industry. The potential damage and wider societal harm that could be caused by large-scale automated decisioning systems is palpable amongst regulators, industry practitioners and consumers~\cite{Recidivism2017,WeaponsMathDestruction2016,PredictAndServe2016}. Two specific concerns that have emerged center on the interpretability and fairness of the decisions resulting from these algorithms. These are not unjustified with cases of unfair decisioning systems manifesting in multiple domains from criminal recidivism~\cite{Recidivism2017} to credit worthiness assessment. In the European Union, these concerns have manifest in the General Data Protection Regulation~\cite{GDPR,GDPRML2017} that enshrines each individual's right to fair and transparent processing. This combined societal and legislative scrutiny has resulted in model interpretability and algorithmic fairness coming to the fore in research~\cite{Dwork2012,FairnessSurvey2019}.

At the broadest level, the concept of algorithmic fairness tackles whether members of specific unprivileged groups are more likely to receive unfavourable decisions from the predictions of a machine learning system. Recent advances have enabled modellers to incorporate fairness at every point of the model building process~\cite{FriedlerComparison,FairnessSurvey2019,Framework2019,DesigningFairAlgos2018}. One embodiment incorporates fairness constraints into the training procedure~\cite{PredRemover2012,MetaLearn2019,FairBoost2020,Goel2018,IBM2018,Nabi2018,FNN2018,FairRed2018,FNN2018}, typically these constraints rely on statistical measures of fairness and are subject to drawbacks~\cite{FairnessImpossibility2018} and trade-offs. These measures rely on \textit{a priori} worldviews and do not incorporate the role of external model auditing or decision explicability in their fairness criteria. This is poorly aligned with how these issues are dealt within industry, where external actors often question the model fairness through building surrogate explanatory models, even if mentally, using the information available to them.

To address these issues, we propose a new definition of fairness we dub ``Fairness by Explicability''. Under this definition, if an external actor's surrogate model cannot produce a \textit{narrative} (i.e., a set of explanations) against the fairness of a particular model, then that particular model can be considered 
\textit{explicably fair}. This definition explicitly frames the perception of an algorithm's fairness as one determined by a combination of an auditor's worldview, data availability, model interpretability framework and measurement/modelling approach. It can be considered complementary to the existing ways of evaluating a model's fairness, since while those may capture risk arising from non-adherence to regulatory requirements, our new ``fairness by explicability'' viewpoint captures the additional and independent risk that may arise from analyses performed by one's own clients~\cite{times}. 

To enforce our ``Fairness by Explicability'' definition, we leverage model interpretability methodologies~\cite{LIME2016,SHAP2017,Hastie2019} to incorporate fairness constraints through adversarial learning. More explicitly, we utilize the SHAP~\cite{SHAP2017,TreeSHAP2019} values of a surrogate adversary model in two ways. The first works by constructing a differentiable fairness regularization term. The second is a modification to the classic AdaBoost algorithm~\cite{AdaBoost1997} to include adversarial attribution values in the weight updates.

We link our fairness approach to statistical fairness~\cite{Metrics2018} via the construction of an appropriate surrogate model. Our approaches are illustrated using a synthetic dataset, the UCI Adult Census Dataset~\cite{uci}, and a commercial credit scoring dataset. These datasets present a diverse evaluation set, with the real-world dataset providing assurance that these approaches are viable in industrial applications. The structure of the papers is as follows: in Section \ref{sec:formulation} we introduce our notation; in Section \ref{sec:SHAP} we provide a brief account of SHAP values and Section \ref{sec:background} discusses statistical fairness measures. Section \ref{sec:fbe} introduces the ``Fairness by Explicability'' worldview and in Section \ref{sec:algos} we present our SHAP-regularized algorithms before discussing the results of the experiments in Section \ref{sec:experiments}. We then state our conclusions and highlight areas of further research in Section \ref{sec:conclusions}.

\section{Notation}
\label{sec:formulation}
To measure the fairness of any algorithm output one needs to define the task objective, the un-/privileged groups to measure fairness against and the favourable outcomes. For the remainder of this paper, we focus on binary classification tasks with a single privileged group indicator $Z$. We denote the other covariates present with $\mathbb{X}$ and the combination of $Z$ with those covariates by $\tilde{\mathbb{X}}$. Furthermore, and without loss of generality, we define the value of $1$ for the target $Y$ and the corresponding model outcomes $\hat{Y}$ as the favourable label. Model outcomes are constructed by applying a threshold to the scores $\bar{Y}$. For each instance $i$, we denote the corresponding values with the appropriate lowercase symbol and subscript, i.e. $y_i$, $z_i$, $x_i$, etc. In this case, $x_i$ and $\tilde{x}_i$ denote vectors and the value of the $j$th covariate is given by $x_{ij}$ and $\tilde{x}_{ij}$.

\section{SHapley Additive Explanations (SHAP)}
\label{sec:SHAP}
SHapley Additive Explanations, or SHAP values~\cite{SHAP2017,TreeSHAP2019}, provide a unified framework for interpreting model predictions. This approach was built off the insight that many other modern explanatory frameworks such as LIME~\cite{LIME2016} and DeepLIFT~\cite{DeepLIFT2017} could be recast as variants of a generic additive feature attribution paradigm. In this paradigm, a simplified explanatory model $\sigma$ is built to explain the original prediction $f$ using simplified binary input vectors $\tilde{x}_{i}' \in \{0, 1\}^M$, where $M$ is the number of features and $i$ is the instance label. These simplified inputs are related to the original feature vectors $\tilde{x}_i$ through the mapping $\tilde{x}_i = h_{\tilde{x}_{i}}(\tilde{x}_{i}')$ and the local explanatory model is given by:
\begin{equation}
 \label{eq:linaddmodel}
 \sigma(\tilde{x}_{i}') = \phi^{i,f}_0 + \sum_{j=1}^{M} \phi^{i,f}_j \tilde{x}_{ij}'.
\end{equation} 
The local feature effect of feature $j$ for model $f$ is $\phi^{i,f}_j$ and global explanations are calculated via the statistics of these values across a dataset. The different explanatory frameworks, e. g. LIME, emerge from specific choices of the mapping function $h_{\tilde{x}_{i}}$, the kernel weighting of instances in the objective ($\pi_{\tilde{x}}$)  and any additional regularization terms $\Omega(\sigma)$ used to fit $\sigma$.  These choices influence the properties of the surrogate model. In Ref.~\cite{SHAP2017}, they showed that only one $\sigma$ satisfies these 3 desirable properties:
\begin{enumerate}
 \item Local Accuracy: $f(\tilde{x}_i)=\sigma(\tilde{x}_{i}')=\sum_{j=0}^{M}\phi^{i,f}_{j}$, when  $\tilde{x}_i=h_{\tilde{x}_{i}}(\tilde{x}_{i}')$.
 \item Missingness: $\tilde{x}_{ij}'=0 \implies \phi^{i,f}_j=0$.
 \item Attribution Consistency: for any two models $f, f'$, the ordering of the differences of the model output when a feature is present vs missing is reflected in their respective attributions of that feature. 
\end{enumerate}
 Its attributions $\phi_j$ are the same Shapley Values first identified in cooperative game theory~\cite{NPersonGame1953,RegressionSHAPLEY2001,SHAPUnique1985,LinearSHAP2014}:
\begin{equation}
 \label{eq:shapattrib}
 \phi^{\bullet,f}_j=\sum_{z' \subseteq \tilde{x}'} \frac{|z'|!(M-|z'|-1)!}{M!}[f_{\tilde{x}}(z')-f_{\tilde{x}}(z'\setminus j)].
\end{equation}
Here, $|z'|$ is the number of non-zero entries in $z'$, $z' \setminus j$ denotes setting the $j$th element of $z'$ to $0$ and the summation is over all $z'$ where the non-zero entries are a subset of the non-zero entries of $\tilde{x}'$. These SHAP values can be estimated for a generic model using KernelSHAP~\cite{SHAP2017} while for specific model families there are efficient computational methods and analytic approximations~\cite{TreeSHAP2019,LinearSHAP2014}. 

\section{Metrics and Statistical Fairness}
\label{sec:background}
To estimate fairness metrics one requires a dataset of $N$ instances with $Y$ and $Z$ as well as the outcomes. Given this data, the appropriate fairness metric is often defined by the worldview(s)~\cite{Worldviews2019} of those auditing the outcomes. These worldviews tend to fall into three broad categories: ``We're all equal''~\cite{barocas2016big}, ``What you see is what you get''~\cite{Dwork2012,roemer2015equality} and causal~\cite{Kilbertus2017,ZhangDirectEffect,Kusner2017,CounterfactualFairGDPR2017,FairBoost2020,Chiappa2018}. The first two categories are statistical in nature and we now discuss their application to the binary task domain.

Statistical fairness metrics relate to the conditional probabilities involving $Y$, $\hat{Y}$ and $Z$. The ``We're all equal'' worldview has numerous group fairness metrics associated with it. These metrics measure any differences in outcome given group membership and seek to balance said outcomes. Contrastingly, ``What you see is what you get'' asserts that the observed data captures the underlying ``truth'' and typically prefers to offer individuals similar outcomes conditional on $Y$. In this work, we consider two of the most common statistical fairness metrics from these categories: ``statistical parity'' difference (SPD) and ``equality of opportunity'' difference (EOD). More formally, these are defined as:
\begin{align}
\label{eq:fairmeasures}
\mathrm{SPD} =|&P(\hat{Y}=1|Z=1) - P(\hat{Y}=1|Z=0)|, \\
\mathrm{EOD} =|&P(\hat{Y}=1|Y=1,Z=1) - \nonumber \\ 
&P(\hat{Y}=1|Y=1,Z=0)|.
\end{align}

Note that a target SPD value can also be calculated by replacing $\hat{Y}$ with $Y$ respectively in Eq.~\ref{eq:fairmeasures}. Both of these measures are estimated from a specified dataset, their value of zero denotes a maximally fair model, and both have trade-offs~\cite{FairnessImpossibility2018} and limitations. For example, SPD can be minimized through randomly modifying outcomes while ignoring all other covariates $\mathbb{X}$ and so can be viewed as a lazy penalization. Contrastingly, minimizing EOD may not reduce any gap in the rate of favourable outcomes between the groups.    

\section{Fairness by Explicability}
\label{sec:fbe}
The traditional statistical fairness metrics presented in Section \ref{sec:background} are not explicitly linked to the domain of model interpretability. Recent work~\cite{MeasUnfairness2020} demonstrated empirically that the SHAP values of $Z$ could capture statistical unfairness provided $Z$ was used as a feature of the model. To formalize an explicit link between model fairness and explicability, we first recall that statistical fairness measures emerge from the worldviews of individuals auditing the model outcomes for fairness. Typically, when trying to understand observations, a human agent (an external actor/auditor) will construct a surrogate model to obtain explanations for their observations. The role of $Z$ in these explanations determines whether the outcomes constructed are perceived as fair or not. Building on this idea, we propose a new worldview to capture the mechanism by which model decisions are evaluated by external actors.

\begin{definition}
\label{def:1}
 Consider a model trained by an auditor to predict $\bar{Y}$ using $Z$ and, optionally, a combination from $\{Y, \mathbb{X}\}$. If this model does not detect any difference in the $Z$ attribution between the $Z={0,1}$ groups, then the predictor model is \textit{explicably fair} with respect to the auditor.
\end{definition}

We dub this worldview ``Fairness by Explicability''. The precise measure of fairness one attains is determined by: the population examined by the auditor, the interpretability framework used, how attributions are calculated and aggregated, and the auditor model developed. This definition can be specialized into a \textit{strong} ``Fairness by Explicability'' form by further requiring that total attribution for $Z$ is also reduced to zero.

Auditors are usually interested in the average attribution of the two groups given a population of data. This informs the metrics used to quantify how ``explicably fair'' a model is. These are:
\begin{align}
 \label{eq:expfairmeasures}
  \mathrm{FE} &= |\frac{\sum_{i, s.t. Z=1} \phi^{i,l}_Z}{N_1} - \frac{\sum_{i, s.t. Z=0} \phi^{i,l}_Z}{N_2}|, \\
  \mathrm{SFE} &= \frac{\sum_{i} |\phi^{i,l}_Z|}{N},
\end{align}
where $\phi^{i,l}_{Z}$ is the SHAP value of $Z$ for instance $i$ for auditor model $l$, $N$ is the total number of instances in the dataset and $N_{1(0)}$ is number of examples when $Z=1(0)$. $\mathrm{FE}$ measures the difference in mean attribution between the two groups. When it is minimized the model is considered fair according to our ``Fairness by Explicability'' definition. The second metric ($\mathrm{SFE}$) measures the total attribution of $Z$ across the population, when minimized the auditor model concludes that the model satisfies the strong version of ``Fairness by Explicability'' and, by definition, the first metric is also zero. These metrics are equivalent to those of Ref.~\cite{MeasUnfairness2020} but in this instance are applied to an external auditor model and are informed by how a typical auditor would aggregate their explicability scores.

From this discussion, ``Fairness by Explicability'' may appear intuitive but difficult to implement and, in general, being ``explicably fair'' does not provide any guarantees of statistical fairness. However, an initial informal connection to the prior fairness worldviews can be made through consideration of specific forms of the auditor models. Intuitively, removing the dependency on $Z$ as measured by an external $l$ will tend to reduce $\bar{Y}$'s dependency on $Z$. This will generally lead to improved SPD and EOD although the decision policy plays a large role in how these two connect. 

\section{Achieving Fairness by Explicability}
\label{sec:algos}
We now present two different approaches for imposing ``Fairness by Explicability'' directly into the training process of gradient-based and adaptive boosting (specifically AdaBoost) algorithms. These approaches rely on inserting a surrogate model $g$ directly into the iterative training procedures.  The form of $g$ is then chosen to account for the examination of an anticipated external auditor whose model is $l$. Both approaches require $Z$ during the training phase only, hence any sensitive attributes defining $Z$ do not need to be supplied at prediction time. In addition to this presentation, we also discuss how the approaches can be linked to the SPD and EOD.

\subsection{SHAPSqueeze}
\label{sec:SHAPSqueeze}
The first approach to imposing ``Fairness by Explicability'' uses a series of differentiable regularizations to penalize unfair attributions. We consider a differentiable loss function of the form:
\begin{equation}
 \label{eq:convex_sum}
 \mathcal{L}_{\mathrm{fair}} = (1-\lambda) * \mathcal{L}_{o} + \lambda * \mathcal{R},
\end{equation}
which we can optimize through gradient-based methods, e.g. stochastic gradient descent. At each iteration, a surrogate model $g$ is fit to the $\bar{Y}$ values. From $g$, the SHAP values of $Z$, and optionally $Y$, are used to calculate the appropriate regularization term ($\mathcal{R}$). In this work, $\mathcal{L}_{o}$ is the binary cross-entropy. Considering the case where $l$ and $g$ are identical, when the associated $\mathcal{R}$ is minimized then the attributions to $Z$ will be zero and \textit{strong} ``Fairness by Explicability'' is satisfied by the model scores $\bar{Y}$. 

The specific form of $g$ we examine is a linear regression model, see the first row of Table~\ref{surrogates}. The SHAP values of interest are given by: 
\begin{equation}
 \label{eq:linearmod}
 \phi^{i,g}_Z = \beta(z_i - \mathbb{E}[Z]).
\end{equation}
Equation (\ref{eq:linearmod}) directly relates the SHAP values of $Z$ to its model coefficient, $\beta$, and the specific realisation of $Z$ for instance $i$. The regularization $\mathcal{R}$ is then simply the sum of the squares of these SHAP values scaled by a constant $C$, see Table~\ref{surrogates}. This constant is used to make the size of the gradients coming from $\mathcal{R}$ and $\mathcal{L}_{o}$ comparable, while $\lambda$ is used to adjust the balance between these two quantities. Moreover, we note that the explicability fairness metrics in Eq.~\ref{eq:expfairmeasures} are proportional to $\beta$  in this instance. Therefore, these specific $g$ and $\mathcal{R}$ will seek to eliminate the linear dependence of the model predictions $\bar{Y}$ on $Z$. Consequently, we expect reductions in the SPD as the model becomes explicably fairer.

To conclude, we note that the use of linear regression makes both the model fitting and SHAP value derivative calculations computationally efficient to perform. However, the approach described is applicable to any $g$ whose SHAP values are differentiable with respect to $\bar{Y}$ and so parametric/kernel regression models could also be employed. In combination with adding more features, this can allow for the consideration of more complex auditors with different worldviews.

\begin{table}
\caption{The surrogate models and regularizations considered in this work.}\label{surrogates}
\centering
\begin{tabular}{|l|l|l|} 
\hline
algorithm & surrogate - $g$ & regularization  \\
\hline
SHAPSqueeze  & \multirow{2}{*}{$\bar{Y} = \beta Z + \alpha$} & $\mathcal{R} = C\sum_i (\phi^{i,g}_Z)^{2}$ \\
SHAPEnforce & & $\mathcal{P}=\begin{cases}
-\phi^{i,g}_Z,\text{if } y_i = 1\\
0,\text{otherwise} 
\end{cases}$ \\
\hline
\end{tabular}
\end{table}

\subsection{SHAPEnforce}
\label{sec:SHAPEnforce} 
The classic AdaBoost algorithm~\cite{AdaBoost1997} trains a model that is a weighted linear combination of weak classifiers. The training process is iterative, with each weak learner ($k_{m}$) being fitted to a reweighted version of the training data. After $R$ iterations, the outputted model is given by $C_{R} = \sum_{m=1}^{R} \alpha_m k_{m}$.  We consider learners that output a score and whose classification output, $\{0, 1\}$, is obtained by thresholding. Traditionally, AdaBoost generates the instance weights for the $m$\textsuperscript{th} training round, $\omega_i^{m}$, by scaling the previous iteration's weights $\omega_i^{(m-1)}$. Instances $k_{m}$ that are incorrectly classified have their weights enhanced by $e^{\alpha_m}$, while correctly classified instances are downweighted by $e^{-\alpha_m}$. As training proceeds, the algorithm increasingly focuses on erroneous examples to improve its predictive performance. 
To incorporate ``Fairness by Explicability'' into AdaBoost, we adjust its reweighting process to consider the SHAP values $\{\phi^{i,g}_j\}, i=1,\dots,N$, of the features $\{j\}$ of a surrogate $g$. This SHAP weighting is introduced through a penalty function ($\mathcal{P}(\{\phi^{i,g}_j\})$) and fairness regularization weight ($\lambda$) which trades off the original weight update with the new penalty.

In effect, this forces weak learners to not only focus on erroneous examples but also those with specific SHAP values as determined by $g$ and $\mathcal{P}$. This pushes the algorithm to improve its predictions on instances with specific SHAP values and is dubbed ``SHAPEnforce''. Furthermore, in contrast to SHAPSqueeze, it is fully non-parametric and only requires that the SHAP values of $g$ can be computed.
\begin{algorithm}
 \KwIn{training examples $\{(x_i, y_i, z_i)\}_{i=1}^{N}$, specification of favourable outcome, a surrogate model $g$, a SHAP penalty function $\mathcal{P}$, and the number of boosting rounds $R$.}
 \KwOut{A classifier $C_{R}(x) = \sum_{m=1}^{R} \alpha_m k_m(x)$.}
 \BlankLine
  Initialize weights $\omega_i^{1} = 1 / N, \forall i$.
 \BlankLine
 \For {$m=1$ \textbf{to} $R$} {
    Fit a weak learner $k_m(x)$ using the training data with weights $\omega_i^{m}$.\;
    Compute the probability of favourable outcome, $\bar{y}_i^m$, and the predicted label $\hat{y}_i^m$ from $k_m$.\;
    Fit $g$ - taking features and the target from $\{(x_i, y_i, z_i, \bar{y}_i^m)\}$.\;
   Compute the $\phi^{i,g}_j$, and the corresponding  weight adjustment $\mathcal{P}(\{\phi^{i,g}_j\})$.\;
    Compute $e_m\leftarrow \mathbb{E}_{\omega^{m}}[\mathbb{1}_{(y\neq k_m(x))}]$.\;
    Compute $\alpha_m = \log((1-e_m)/e_m)$.\;
    Update the instance weights: $\omega_i^{(m+1)}\leftarrow \omega_i^m [(1-\lambda) * e^{\alpha_m (\mathbb{1}_{(y\neq k_m(x))} - \mathbb{1}_{(y = k_m(x))})} + \lambda e^{\mathcal{P}(\{\phi^{i,g}_j\})}]$.\;
    Set $\omega^{m+1}_i \leftarrow \frac{\omega^{m+1}_i}{\sum_i \omega^{m+1}_i}$.
 }
 \caption{SHAPEnforce}
 \label{alg:SHAPENFORCE}
\end{algorithm}

Algorithm~\ref{alg:SHAPENFORCE} presents the pseudo-code for ``SHAPEnforce''. The learning approach can be qualitatively interpreted as a two-player game. Expanding on this view, at each stage the predictive learner makes a move by constructing a weak learner and attempts to reweight the training data as-if the surrogate had not acted up to that point. Similarly, once the learner is constructed the surrogate acts to reweight the dataset in its own best interest. The regularization weight $\lambda$ then controls the resulting outcome between these two competing actions. 

In this work, we consider a linear surrogate model trained on data where $Y=1$ whose form and associated $\mathcal{P}$ is shown in Table~\ref{surrogates}. We again approximate the SHAP values using Eq.~\ref{eq:linearmod}. The $\mathcal{P}$ considered is local in nature and, conditioned on $Y=1$, will downweight any examples with positive SHAP values while upweighting those with negative values. This forces the predictor model to focus on instances where the $Z$ attributions have a negative impact on the favourable outcome and where the weak learner has made mistakes when the target is favourable, i.e. $Y=1$. By focusing on the examples with negative $Z$ attribution, their $Z$ attribution will be increased at the next round, hence the explicability fairness, as determined by an equivalent $l$, will tend to increase. This choice of $\mathcal{P}$ further reflects the intuition that unprivileged groups are likely to have unfavourable predictions from weak learners and hence negative $Z$ attribution. Furthermore, with the focus on examples where $Y=1$ we expect this modification to reduce the EOD. Finally, $\mathcal{P}$ is related to a fairness regularization term previously applied to neural networks~\cite{Beutel2019}. Our work formalizes this previously ad-hoc loss as a ``Fairness by Explicability'' regularizer with an appropriate auditor.

\begin{figure}
  \centering
  \includegraphics[width=\linewidth]{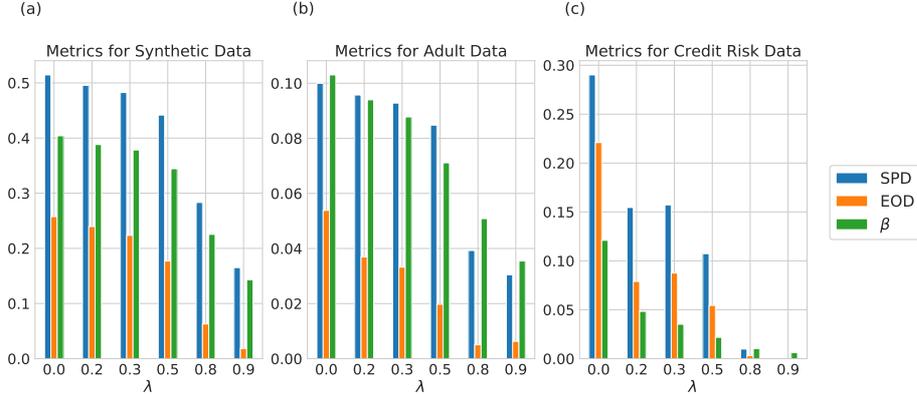}
  \caption{Fairness metrics for SHAPSqueeze plotted with varying regularization strength for the (a) synthetic, (b) Adult and (c) Credit Risk test datasets. We set $C=1$ for the synthetic data, $C=10$ for Adult and $C=100$ for the Credit Risk evaluations.}
\label{fig:results_shapsqueeze}
\end{figure}

\section{Computational Experiments}
\label{sec:experiments}
To evaluate our algorithms we consider three binary classification datasets: a synthetic dataset, the UCI Adult dataset~\cite{uci}, and a commercial Credit Risk dataset. The train/test splits are shown in Table~\ref{datasets}. The datasets were preprocessed so categorical variables were one-hot encoded and numeric variables were converted to their standard score. 

We exemplify the SHAPSqueeze objectives using XGBoost~\cite{XGBoost2016}. In each experiment, we evaluate the algorithms predictive performance, as measured by accuracy/precision and ROC AUC, as well as measuring the SPD and EOD. To determine these quantities, we use a fixed threshold policy. For SHAPSqueeze, in the case of the synthetic and UCI Adult dataset, this threshold is $0.5$ while a more risk-averse threshold of $0.85$ is set for the commercial Credit Risk dataset. This higher threshold better reflects real-world business practices in this domain. SHAPEnforce, being a modification to AdaBoost, is less calibrated than the SHAPSqueeze implementation and so a threshold of $0.5$ is used in all cases. Additionally, we build linear regression auditor models on the test set to measure the explicability fairness. The equations defining $l$ are the same as the $g$ employed, and so the explicability fairness is given by the coefficient $\beta$ of the fitted $l$, see Table~\ref{surrogates}. Note for SHAPEnforce, $l$ is built on the data subset where $Y=1$.

\begin{table}
\caption{Datasets used for the algorithm evaluation.}\label{datasets}
\centering
\begin{tabular}{|l|l|l|} 
\hline
dataset & train size & test size  \\
\hline
Synthetic & 75000 & 25000\\
\hline
Adult & 32561 & 16281\\
\hline
Credit Risk & 48112 & 23697\\
\hline
\end{tabular}
\end{table}

\subsection{Datasets}
\label{sec:datasets}
\subsubsection{Synthetic Data}The synthetic dataset was generated to exhibit a very large SPD. To construct this, the distribution of $\mathbb{X}$ is conditional on $Z$ and $Y$ is determined by $Z$ and $\mathbb{X}$. Specifically, $Z$ was sampled from a Bernoulli distribution and $\mathbb{X}$ contains three sets of covariates: ``safe'' covariates $\mathbb{X}_s \sim \mathcal{N}(0, 1)$, ``proxy'' covariates ($\mathbb{X}_p$) and ``indirect effect'' covariates ($\mathbb{X}_i$). The latter two are sampled from $\mathcal{N}(Z, 1)$. From this, the log-odds of the binary target ($S_Y$) are given by $0.25 \mathbf{w} \cdot(\mathbb{X}_i + \mathbb{X}_s) + 1.25 Z$, $\mathbf{w}$ is a vector of ones. The target $Y$ is then sampled from $\mathrm{Bern} \left( \frac{1}{1 + \mathrm{e}^{-S_Y}} \right)$. Using this approach we sampled a dataset with $10$ safe, $4$ indirect effect and $2$ proxy variables. Furthermore, the sampled dataset was such that approximately $90\%$ of the favourable outcomes were obtained by the privileged group.

\subsubsection{Adult Census}
The goal is to predict whether a person will have an income below or above $\$50\mathrm{k}$. In this dataset, we consider the variable sex as our protected attribute and removed race, marital status, native country and relationship from our models. The other covariates measure financial information, occupation and education.

\subsubsection{Private Credit Risk Dataset}
In this dataset, we are trying to infer a customer's default probability given curated information on their current account transactions. We are interested in removing bias related to age. We binarize the age variable dividing our examples in two groups, an ``older'' (unprivileged) group of people over 50 and a ``younger'' group of people under 50 years old.

\subsection{Results}
\begin{figure}
  \centering
  \includegraphics[width=\linewidth]{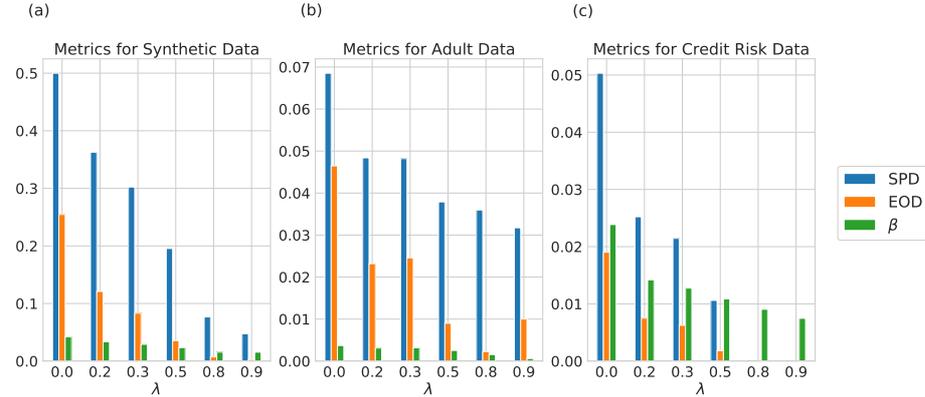}
  \caption{Fairness metrics for SHAPEnforce, using the penalty $\mathcal{P}$ in Table~\ref{surrogates}, plotted with varying regularization strength. Results for the synthetic, Adult and Credit Risk test datasets are shown in (a), (b) and (c) respectively.}
\label{fig:results_shapenforce}
\end{figure}
Results for SHAPSqueeze on the test datasets are shown in Fig.~\ref{fig:results_shapsqueeze}. We observe that across all 3 datasets increasing $\lambda$ induces fairness as observed by reductions in SPD, EOD and $\beta$. We set $C=1$ for the synthetic dataset, $C=10$ for Adult and for the Credit Risk dataset we set $C=100$. These values were chosen to ensure the mean gradients from $\mathcal{L}_o$ and $\mathcal{R}$ in the intermediate stages of training, i.e. $\sim 100$ iterations, when $\lambda=0.5$ were on the same order of magnitude and effective. For the synthetic data, we observe $\beta$ drops from $0.404$ at $\lambda=0$ to $0.142$ at $\lambda=0.9$. It is accompanied by a tolerable drop in the AUC and accuracy of $0.04$ in both cases. Similarly, the SPD is reduced by roughly $0.35$ while the EOD is almost eliminated, taking a value of $0.018$ at $\lambda=0.9$. Increasing $\lambda$ further, $\beta$ approaches zero and is faithful to our \textit{strong} ``Fairness by Explicability'' definition.

We observe the same patterns for the fairness metrics when SHAPSqueeze is applied to the Adult and Credit Risk datasets. In the former, we observe a reduction of roughly $0.04$ in accuracy and AUC with increasing $\lambda$, at $\lambda=0.9$ these take values of $0.80$ and $0.83$ respectively. In the latter case, the precision is reduced by $\approx0.12$ and the AUC drops by $\approx0.07$ as we change $\lambda$ from $0$ to $0.9$. Contrastingly, for Adult, we observe an increase in precision (from $0.76$ to $0.98$) as the fairness regularization increases the scores beyond the classification threshold. A similar effect is seen in the Credit Risk dataset where we observed an increase in the accuracy from $0.744$ to $0.837$ as $\lambda$ was increased to $0.9$. This increased accuracy is attributed to the conservative threshold of $0.85$ employed. This threshold also results in the SPD and EOD being eliminated at $\lambda=0.9$ as the regularization pushes all of the scores above $0.85$. At this point $\beta$ is roughly $0.006$ demonstrating that even when the SPD and EOD are zero a model may not be $100\%$ explicably fair. This highlights the differences in fairness definition and, in particular, the use of $\bar{Y}$ and not $\hat{Y}$ when measuring explicable fairness. To avoid this scenario one would either reduce $C$ or select a different $\lambda$ value. At $\lambda=0.7$, the model has SPD, EOD and $\beta$ values of $0.035$, $0.013$ and $0.014$ respectively. It is also performant with tolerable drops in the AUC ($0.06$) and precision ($0.1$) observed. 

The results for SHAPEnforce are presented in Fig.~\ref{fig:results_shapenforce}. In all cases, we observe the EOD, SPD, AUC and accuracy decrease with increasing $\lambda$. For the synthetic data, the accuracy drops by approximately $0.08$ from $0.828$ to $0.75$ as we increase $\lambda$. This is accompanied by a drop of $\approx 0.03$ in the AUC from $0.868$ to $0.836$ as we change $\lambda$ from $0$ to $0.9$. Compared to the statistical fairness metrics, we observe smaller improvements in the explicable fairness. Furthermore, the decreasing trend of $\beta$ is less pronounced and consistent compared to SHAPSqueeze. This was expected for two reasons. Firstly, the unregularized AdaBoost model is explicably fairer than XGBoost and so there is less explicable unfairness to remove. Secondly, we expected the heuristic nature of the modification provides no guarantees on explicable fairness and so the magnitude of the reduction is not guaranteed. For the synthetic dataset we observe a decrease in $\beta$ of $\approx 63\%$ as we increase $\lambda$ from $0$ to $0.9$. Moving to the Adult results, we observe $\beta$ decreases by approximately $84\%$ on changing $\lambda$ from $0$ to $0.9$. The SPD and EOD are reduced to $0.03$ and $0.01$ respectively with tolerable drops in accuracy ($0.02$) and AUC ($0.01$) observed. For the Credit Risk data, we again observe explicable fairness improvements, on the order of $69\%$ as we increase $\lambda$. This is accompanied with the SPD and EOD being eliminated for $\lambda > 0.6$. Similar to SHAPSqueeze, this elimination is due to the regularization pushing all scores below the threshold for $\lambda > 0.6$. In practice one would use a model from another $\lambda$, such as $\lambda=0.2$, where the SPD and EOD are reduced by roughly $50\%$ and $61\%$ respectively while the precision and AUC take values of $0.85$ and $0.84$ respectively. This represents a drop of $\approx0.01$ for the former while the latter is consistent with the unregularized model. However, at this point, $\beta$ is only reduced by approximately $40.5\%$ compared to $\lambda=0$.

\section{Conclusions}
\label{sec:conclusions}
In this work, we developed a novel fairness definition, ``Fairness by Explicability'', that gives the explanations of an auditor's surrogate model primacy when determining model fairness. We demonstrated how to incorporate this definition into model training using adversarial learning with surrogate models and SHAP values. This approach was implemented through appropriate regularization terms and a bespoke adaptation of AdaBoost. We exemplified these approaches on 3 datasets, using XGBoost in combination with our regularizations, and connected our choices of surrogate model to ``statistical parity'' and ``equality of opportunity'' difference. In all cases, the models trained were explicably and statistically fairer, yet still performant. This methodology can be readily extended to other interpretability frameworks, such as LIME~\cite{LIME2016}, with the only constraint being that $\mathcal{R}$ must be appropriately differentiable. Future work will explore more complex surrogate models and different explicability scores in the proposed framework.

\section{Related Work}
In recent years, there has been significant work done in both model interpretability, adversarial learning and fairness constrained machine learning model training.

\textit{Interpretability}: Ref.~\cite{SHAP2017} provided a unified framework for interpreting model predictions. This framework unified several existing frameworks, e.g. LIME~\cite{LIME2016} and DeepLift~\cite{DeepLIFT2017}, and it can be argued to be the ``gold standard'' for model interpretability. It provides both local and global measures of feature attribution and through the KernelSHAP algorithm, is model agnostic. Further work has introduced computationally efficient approximations to the SHAP values of \cite{SHAP2017} for tree-based models~\cite{TreeSHAP2019}. Other works in interpretability have focussed on causality for model interpretability. These approaches provide insight into \textit{why} the decision was made, rather than an explanation of the model predictive accuracy and are frequently qualitative in nature. Ref.~\cite{CFE2020} is a recent exception, where the counterfactual examples generated obey realistic constraints to ensure practical use and are examined quantitatively through bespoke metrics. 

\textit{Adversarial Training}: Ref.~\cite{Beutel2017} used adversarial training to remove EOD while a framework for learning adversarially fair representations was developed in Ref.~\cite{madras18a}. Similar, in Ref.~\cite{IBM2018} an adversarial network~\cite{GANs} was used to debias a predictor network, their specific approach compared favourably to the approach of~\cite{Beutel2017}. 

\textit{Training Fair Models}: typically, fair learning methodologies have tended to focus on incorporating statistical fairness constraints directly into the model objective function. Ref.~\cite{FNN2018} combined neural networks with statistical fairness regularizations but their form restricts their applicability to neural networks. Similarly, Ref.~\cite{Goel2018} trained a fair logistic regression using convex regularizations and addresses proportionally fair classification. Other works have viewed fair model training as one of constrained optimization~\cite{Zafar2017a,Zafar2017c} or have created meta-algorithms for fair classification~\cite{MetaLearn2019}. 

In these works, the approaches to fair learning have tended to focus on fairness metrics associated with more traditional worldviews and less focus on model explicability. Similarly, the role of model explicability in fairness, to the authors' knowledge, has not been used directly in fair model training but instead research has focussed on the consistency and transparency of explanations. Our work is novel as it places the role of model explicability at the core of a new fairness definition and develops an adversarial learning methodology that is applicable to adaptive boosting and any model trained via gradient-based optimization. In the former case, our proposed algorithm is fully non-parametric where the adversary can come from any model family provided the corresponding explicability scores, in this case SHAP values, can be computed.

\bibliographystyle{splncs04}

\end{document}